\documentclass[sigconf, screen, nonacm]{acmart}  %arxiv

\usepackage[ruled,vlined]{algorithm2e} % 必须添加的宏包

\usepackage{tabularx,booktabs,makecell}  % 添加 makecell 包处理换行
\usepackage{graphicx}                   % 调整字体尺寸
\usepackage{pifont}                     % 保留原有符号

\usepackage{tabularx}
\usepackage{multirow}
\usepackage[table]{xcolor}
\usepackage{makecell}

\usepackage{array}
\usepackage{multirow}

\usepackage{amsmath}    % 提供 \bm 基础支持
\usepackage{bm}         % 增强的粗体符号（可选但推荐）

\usepackage{mathtools}  % 提供 \mathclap 命令
\usepackage{amsfonts}

\newcolumntype{Y}{>{\raggedright\arraybackslash}X}  % 自定义列格式
\AtBeginDocument{%
  \providecommand\BibTeX{{%
    \normalfont B\kern-0.5em{\scshape i\kern-0.25em b}\kern-0.8em\TeX}}}

\begin{document}

\title{Adaptive Spiking Neurons for Vision and Language Modeling}

% \affiliation{%

\author{Chenlin Zhou}
\affiliation{%
  \department{School of Electronic and Computer Engineering, Peking University}
  % \institution{Peking University}
  \city{Shenzhen}
  \country{China}
}
% \email{zhouchenlin25@stu.pku.edu.cn}

\author{Sihang Guo}
\affiliation{%
  % \department{School of Computer Science}
  \institution{Harbin Institute of Technology}
  \city{Shenzhen}
  \country{China}
}
% \email{zhouchenlin25@stu.pku.edu.cn}

\author{Jiaqi Wang}
\affiliation{%
%   \department{School of Computer Science}
  \institution{Harbin Institute of Technology}
  \city{Shenzhen}
  \country{China}
}
% \email{zhouchenlin25@stu.pku.edu.cn}

\author{Dongyang Ma}
\affiliation{%
  \department{School of Computer Science}
  \institution{Peking University}
  \city{Beijing}
  \country{China}
}

\author{Jin Cheng}
\affiliation{%
  \department{School of Electronic and Computer Engineering, Peking University}
  % \institution{Peking University}
  \city{Beijing}
  \country{China}
}

\author{Qingyan Meng}
\affiliation{%
  % \department{School of Computer Science}
  \institution{Pengcheng Laboratory}
  \city{Shenzhen}
  \country{China}
}
% \email{zhouchenlin25@stu.pku.edu.cn}

\author{Zhengyu Ma}
\affiliation{%
  % \department{School of Computer Science}
  \institution{Pengcheng Laboratory}
  \city{Shenzhen}
  \country{China}
}
% \email{zhouchenlin25@stu.pku.edu.cn}

\author{Yonghong Tian}
\affiliation{%
  \department{School of Computer Science}
  \institution{Peking University}
  \city{Beijing}
  \country{China}
}
% \email{zhouchenlin25@stu.pku.edu.cn}

% \author{Anonymous Authors}
%% You do not have to enter your paper ID

% \acmSubmissionID{6403}

\begin{abstract}
Regarded as the third generation of neural networks, Spiking Neural Networks (SNNs) have garnered significant traction due to their biological plausibility and energy efficiency.
Recent advancements in large models necessitate spiking neurons capable of high performance, adaptability, and training efficiency.
In this work, we first propose a novel functional perspective that provides general guidance for designing the new generation of spiking neurons.
Following the insightful guidelines, we propose the Adaptive Spiking Neuron (ASN), which incorporates trainable parameters to learn membrane potential dynamics and enable adaptive firing. ASN adopts an integer training and spike inference paradigm, facilitating efficient SNN training. To further enhance robustness, we propose a specialized variant of ASN, the Normalized Adaptive Spiking Neuron (NASN), which integrates normalization to stabilize training.
We evaluate our neuron model on 19 datasets spanning five distinct tasks in both vision and language modalities, demonstrating the effectiveness and versatility of the ASN family. Our ASN family is expected to become the new generation of general-purpose spiking neurons.
%To enhance training stability in complex SNN architectures, we incorporate a normalization mechanism into ASN, resulting in a variant termed Normalized ASN (NASN). Furthermore, to improve neuronal expressiveness in event-driven computation, we introduce ternary spike coding, leading to the Ternary ASN (TASN).
\end{abstract}

% \ccsdesc[500]{Computing methodologies~Computer vision}

%    <concept>
%        <concept_id>10010147.10010178.10010224.10010245.10010250</concept_id>
%        <concept_desc>Computing methodologies~Multimedia and Language</concept_desc>
%        <concept_significance>500</concept_significance>
%    </concept>
%  </ccs2012>

% \ccsdesc[500]{Computing methodologies~Multimedia and Language, Multimedia Applications}



\maketitle

\section{Introduction}

Spiking Neural Networks (SNNs), widely regarded as the third generation of neural networks \citep{maass1997networks}, have attracted increasing attention due to their high biological plausibility and superior energy efficiency. Unlike conventional Artificial Neural Networks (ANNs), which rely on continuous-valued activations, SNNs communicate through discrete spikes and process information via event-driven dynamics. This spike-based computation paradigm enables neurons to remain inactive in the absence of events, resulting in sparse computation. Accordingly, SNNs replace traditional high-power dense multiply–accumulate (MAC) operations with sparse low-power accumulate (AC) operations, thereby achieving substantial energy savings \citep{roy2019towards}. SNNs are increasingly being considered a promising paradigm for energy-efficient artificial intelligence \citep{fang2023SpikingJelly}.

% Spiking neurons are the key components of SNNs, and the increasing complexity of modern artificial intelligence tasks has raised higher demands on their efficiency and adaptability. 
The pivotal role of spiking neurons within SNNs means that, as artificial intelligence (AI) tasks grow in complexity, the design of these neural units must evolve to meet higher benchmarks of efficiency and functional adaptability.
The comparison of the previous foundation spiking neurons in the SNN domain is shown in the Table \ref{neuron_complexity}.
In addition to training efficiency, architectural compatibility, and spike-driven properties, adaptive firing with respect to the membrane potential constitutes another key property. As an essential characteristic of general-purpose spiking neurons, this adaptability regulates firing dynamics to ensure effective information transmission, preventing both excessive and insufficient neuronal activity.
%
% Among previous foundation neuron models, which are shown in the Table \ref{neuron_complexity}, 
%
The Leaky Integrate-and-Fire (LIF) neuron is hindered by low training efficiency and a lack of membrane potential adaptivity.
The Parametric Leaky Integrate-and-Fire (PLIF) neuron \citep{fang2021incorporating} introduces learnable membrane time constants to capture dynamic neuronal behavior, enabling adaptive firing. However, PLIF still suffers from low training efficiency, which is further exacerbated by the sigmoid-constrained learnable parameters.
Parallel Spiking Neuron (PSN) enhances adaptivity via parameter matrices and improves training efficiency; nevertheless, it retains T-dimensional training, leading to non-negligible computational time and memory overheads. Furthermore, the compatibility of PSN with mainstream spiking backbones, such as Spiking Transformers \citep{zhou2023spikformer,zhou2026spikingformer,yao2023spike}, remains an open challenge, likely attributable to optimization difficulties associated with its parameter matrices. 
On a different trajectory, the Integer LIF (ILIF) pioneers an efficient integer Training and Spike Inference paradigm. Building on this, NILIF \citep{lei2025spike2former} incorporates normalization to enhance neuronal training stability.
However, both ILIF and NILIF are constrained by insufficient membrane potential adaptivity, limiting their potential as a universal neuron. 
Although the recent SpikingBrain \citep{pan2025spikingbrain} enhances ILIF by employing an adaptive threshold derived from pre-calculated average membrane potentials, this input-dependent strategy remains incongruent with the inherently spike-driven paradigm essential for efficient neuromorphic computation.

%$\bigcirc$
\begin{table*}[htb]
\begin{center}
\caption{Comparison with other foundation spiking neurons in the SNN domain, including LIF \cite{fang2021deep}, PLIF \cite{fang2021incorporating}, PSN \cite{fang2023parallel}, ILIF \cite{yao2023spike}, and NILIF\citep{lei2025spike2former}.  \ding{109} indicates that PSN is substantially faster than LIF and PLIF, while remaining notably slower than ILIF and NILIF in training. Our ASN family (including ASN and NASN) satisfies the four characteristics: Efficient Training, Adaptive Firing, Architecture Compatibility, Spike-Driven Inference.}
\label{neuron_complexity}
\begin{tabular}{lp{3.0cm}<{\centering}p{3.0cm}<{\centering}p{3.5cm}<{\centering}p{3.0cm}<{\centering}} 
\toprule
{Spiking Neurons} & Efficient Training & Adaptive Firing & Architecture Compatibility  & Spike-Driven Inference \\
\midrule
LIF\citep{gerstner2014neuronal} & \ding{55} & \ding{55} & \ding{51} &  \ding{51} \\ 
PLIF\citep{fang2021incorporating} & \ding{55} & \ding{51} & \ding{51} &  \ding{51} \\ 
PSN\citep{fang2023parallel} &  \ding{109}  & \ding{51} & \ding{55} &  \ding{51} \\ 
ILIF\citep{luo2024integer} & \ding{51} & \ding{55} & \ding{51} &  \ding{51} \\ 
NILIF\citep{lei2025spike2former} & \ding{51} & \ding{55} & \ding{51} &  \ding{51} \\ 
\midrule
\textbf{ASN, NASN}  & \ding{51} & \ding{51} & \ding{51} &  \ding{51} \\  
\bottomrule
\end{tabular}
\end{center}
% \vspace{-0.3cm}
% \vspace{-0.2cm}
\end{table*}

% The landscape of modern artificial intelligence is increasingly characterized by multi-tasking and large-scale parameterization, yet conventional spiking neurons often encounter significant bottlenecks under these demands. To address this limitation, 

Based on the above analysis, we adopt a novel functional perspective to systematically revisit the fundamental design principles of the new generation of spiking neurons in SNNs.
We argue that the four basic characteristics of spiking neurons—efficient training, adaptive firing, architecture compatibility, and spike-driven inference—must be considered simultaneously.
Guided by these principles, we propose the Adaptive Spiking Neuron (ASN) and its variant, the Normalized Adaptive Spiking Neuron (NASN). Both ASN and NASN are general spike-driven neurons that combine efficient training with adaptive membrane potential dynamics.  
The ASN family (ASN and NASN) introduces a layer-wise learnable parameter \(\alpha\) to dynamically shift the activation window in the ILIF family (ILIF and NILIF), aligning with the evolving distribution of membrane potentials across layers. This adaptive mechanism enhances both representational capacity and distributional flexibility, making ASN and NASN particularly suitable for vision and language modeling.
Overall, this work advances the design of spiking neurons in SNNs. Our main contributions are summarized as follows.

% 1) We systematically analyze the limitations of existing foundation spiking neurons and provide a functional view for the new generation of spiking neuron design.

% 2) Based on our insightful guidelines, we propose the Adaptive Spiking Neuron (ASN) and its variants, the Normalized Adaptive Spiking Neuron (NASN).  Both ASN and NASN are general spike-driven neurons, enabling both efficient training and adaptive membrane potential behavior.

% 3) We conduct extensive evaluations across 19 datasets spanning five distinct tasks in both vision and language modalities. Our method demonstrates consistently competitive performance, highlighting its effectiveness and strong generalizability.

\begin{itemize}
    \item We systematically analyze the limitations of existing foundation spiking neurons and provide a functional view for the new generation of spiking neuron design.
    
    \item Based on our insightful guidelines, we propose the Adaptive Spiking Neuron (ASN) and its variants, the Normalized Adaptive Spiking Neuron (NASN).  Both ASN and NASN are general spike-driven neurons, enabling both efficient training and adaptive membrane potential behavior.
    
    \item We conduct extensive evaluations across 19 datasets spanning five distinct tasks in both vision and language modalities. Our method demonstrates consistently competitive performance, highlighting its effectiveness and strong generalizability.
\end{itemize}

\section{Related works}
Spiking neural networks are mainly obtained through two ways: ANN-to-SNN conversion \citep{cao2015spiking, rueckauer2017conversion, wang2022signed} and Direct Training\citep{fang2021deep,zhou2023spikformer,zhou2026spikingformer,zhou2024direct}. This work mainly discusses the latter way.

% \subsection{SNNs in Vision and Language Tasks}
\subsection{Spiking Neural Networks in Vision Tasks}
% Spiking Neural Networks (SNNs) have been increasingly applied to visual tasks 
% \citep{zhou2023spikformer, zhou2023spikingformer, yao2023spike, yao2024spike, zhou2024qkformer, yao2025scaling, fangspiking}. 
% Spikformer \citep{zhou2023spikformer} introduced {Spiking Self-Attention (SSA)}, replacing softmax with sparse spike-form Query, Key, and Value, and achieved 74.81\% accuracy on ImageNet-1k with only four time steps, demonstrating the potential of transformer-based SNNs. 
% %
% Building on this, Spikingformer \citep{zhou2026spikingformer} employed a pre-activation shortcut to eliminate floating-point multiplications and reduce firing rates, further improving accuracy to 77.64$\%$. Serving as a backbone, Spikingformer combines spike-driven computation with global modeling capabilities.  

Spiking Neural Networks (SNNs) have gained increasing traction in visual tasks \citep{zhou2023spikformer, zhou2026spikingformer, yao2023spike, yao2024spike, zhou2024qkformer, yao2025scaling, fangspiking}. 
A notable advancement is Spikformer \citep{zhou2023spikformer}, which introduces Spiking Self-Attention (SSA) by replacing softmax with sparse spike-form Query, Key, and Value, achieving 74.81$\%$ accuracy on ImageNet-1k with only four time steps—highlighting the potential of transformer-based SNNs. 
Building on this, Spikingformer \citep{zhou2026spikingformer} incorporates a pre-activation shortcut to eliminate floating-point multiplications and reduce firing rates, boosting accuracy to 77.64$\%$. As a backbone architecture, it effectively unifies spike-driven computation with global modeling capabilities.
Spike-Driven Transformer \citep{yao2023spike} proposed {Spike-Driven Self-Attention (SDSA)}, which relies solely on masking and addition, achieving 77.07\% on ImageNet-1k with substantially lower computational cost. More recently, hierarchical visual spiking transformers 
\citep{yao2024spike, zhou2024qkformer, yao2025scaling, fangspiking} have surpassed 80\% accuracy on ImageNet while maintaining high energy efficiency.  
% For example, MaxFormer \citep{fangspiking} observes that spiking neurons tend to suppress certain frequency components during information transmission. To mitigate this effect, MaxFormer applies max pooling in the patch embedding stage to recover high-frequency components.

\subsection{Spiking Neural Networks in Language Tasks}
SpikeBert \citep{lv2023spikebert} extends Spikformer \citep{zhou2023spikformer} to language modeling by introducing a softmax-free spiking transformer architecture, along with a two-stage knowledge distillation strategy. Specifically, it first performs pretraining by distilling knowledge from BERT on large-scale unlabeled corpora, followed by task-specific fine-tuning using a BERT model trained on the same downstream data. 
% With this approach, SpikeBert achieves 59.7$\%$ accuracy on the GLUE development benchmark \citep{wang2018glue}.
%
SpikeGPT \citep{zhu2023spikegpt} builds a spiking language model upon the RWKV architecture \citep{peng2023rwkv}.
% , while still retaining exponential and division operations analogous to those used in softmax.
%
SpikeLM \citep{xing2024spikelm} introduces a spike-based formulation with bidirectional ternary firing, yielding competitive performance in language modeling. 
%
% However, it largely preserves key elements of conventional attention mechanisms, including floating-point matrix multiplications, softmax operations, and non-spiking activations (e.g., GeLU \citep{hendrycks2016gaussian}) within the MLP blocks.
%
SpikeLLM \citep{xing2024spikellm} incorporates generalized integrate-and-fire (GIF) neurons within an optimal brain spiking framework for transformer architectures. 
% Nevertheless, it remains closely aligned with LLaMA \citep{touvron2023llama}, retaining several non-spiking and nonlinear components, such as softmax-based attention, SiLU activations in MLP layers, and rotary positional embeddings \citep{su2024roformer}, all of which involve non-spike-driven computations in the query and key projections.
However, both SpikeLM and SpikeLLM retaining several non-spiking and nonlinear components, such as softmax-based attention, non-spiking activations in MLP layers.
\citet{zhou2026winnertakeallspikingtransformerlanguage} incorporates the Winner-Take-All (WTA) biological mechanism into spiking transformers and proposes the WTA-based Encoder-only Spiking Transformer (WE-SpikingFormer) for masked language modeling, as well as the WTA-based Decoder-only Spiking Transformer (WD-SpikingFormer) for causal language modeling, systematically exploring softmax-free, spike-driven transformers trained directly for language tasks.

\begin{figure*}[tb]
\centering
% \includegraphics[
%         width=\textwidth,        % 或者指定比例
%         trim=0 335 0 350, 
%         clip
%     ]{./figs/PI-LIF.pdf}
\includegraphics[width=0.85\textwidth]{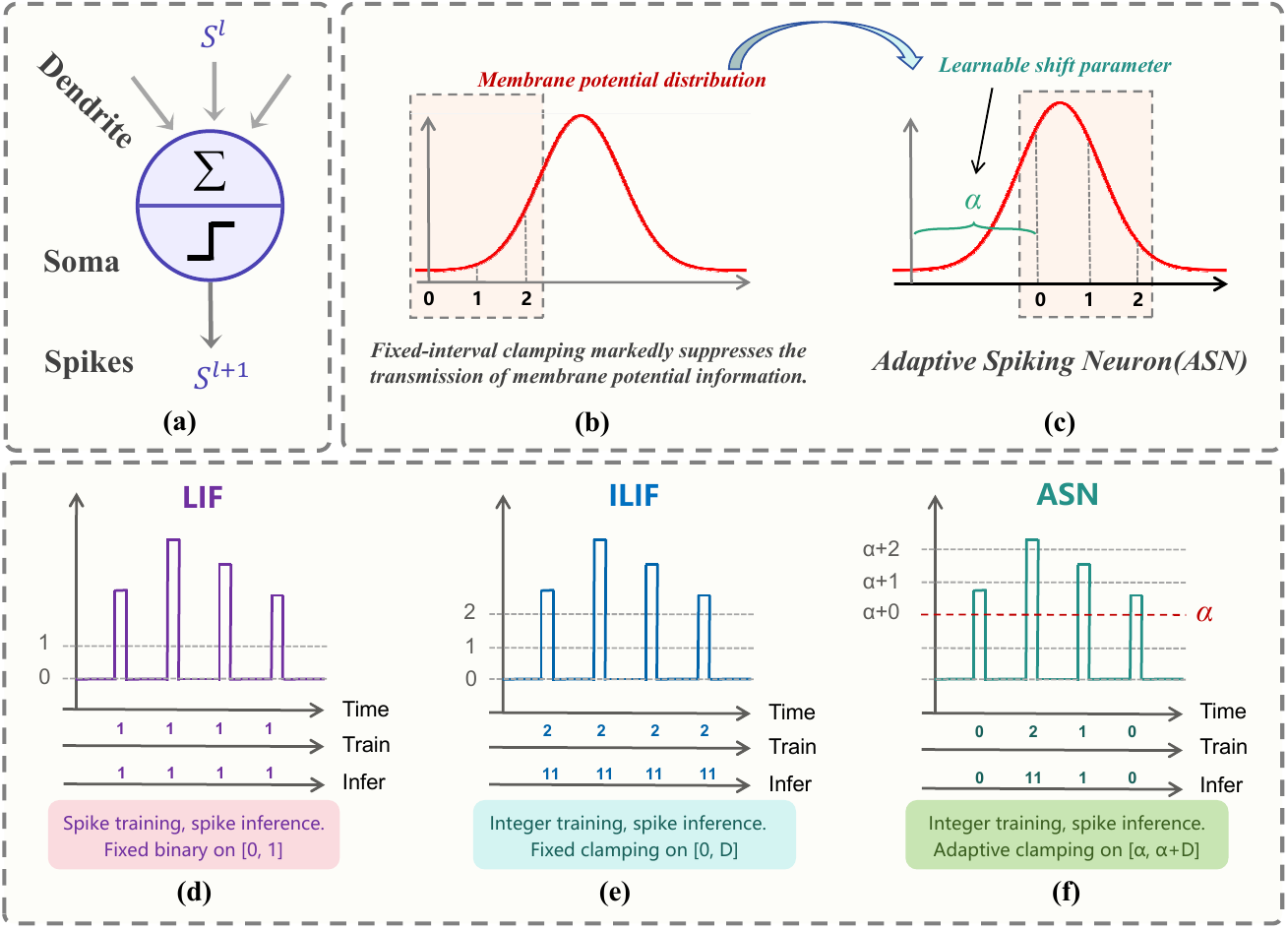}
\caption{Overview of Adaptive Spiking Neuron(ASN).
(a)  schematics of a biological spiking neuron. 
% From the perspective of the neuron group level.
(b) depicts a neuron lacking membrane potential adaptation, thereby suppressing the transmission of membrane potential information. 
(c) illustrates the proposed neuron, where a learnable shift parameter enables adaptive firing, thereby ensuring better membrane potential information transmission.
From the perspective of spike dynamics at the neural level.
(d) and (e) show the dynamics of LIF (spike training and spike inference) and ILIF (integer training and spike inference) neurons, respectively. Both employ fixed-interval clamping, resulting in significant membrane potential quantization bias. 
% (d) shows the LIF neuron with spike training and spike inference. 
% (e) shows ILIF neuron with Integer training and spike inference. Both (d) and (e) employ fixed-interval clamping, resulting in a significant bias in membrane potential quantization.
(f) shows the spike dynamics of our proposed ASN with adaptive firing, achieving more precise spike quantization of membrane potential.
}
\label{main_fig}
\end{figure*}

\subsection{Foundation Spiking Neurons}
% The Leaky Integrate-and-Fire (LIF) \cite{gerstner2014neuronal} model is the most commonly used neuron in SNNs due to its simplicity. Meanwhile, many improved versions of neurons based on LIF have recently emerged to adapt to modern artificial intelligence tasks.
% %
% The Parametric Leaky Integrated-and-Fire (PLIF) spiking neuron \citep{fang2021incorporating} incorporates learnable membrane time constants to learn neuronal dynamics.
% %
% PSN \citep{fang2023parallel} introduces a learnable parameter matrix and removes the reset of LIF to generate hidden states that are independent of their predecessors, resulting in parallelizable neuronal dynamics.
% %
% LIF, PLIF, and PSN are representative approaches within the Spike Training and Spike Inference (STSI) paradigm.
% %
% Another direct training paradigm for SNNs is Integer-Valued Training and Spike Inference (ITSI).
% ILIF \citep{luo2024integer} proposes this paradigm, which employs integer-valued activations during training while maintaining spike-driven inference by extending virtual timesteps.
% %
% NILIF\citep{lei2025spike2former} proposed the normalized version of ILIF to solve the training stability problem of SNNs with some complex architectures.
% %
% SpikingBrain\citep{pan2025spikingbrain} adopts an adaptive threshold in ILIF by pre-calculating the average membrane potential value based on the input. However, this input-dependent adaptation is not well-suited to the spike-driven nature of neuromorphic computation.

The LIF \citep{gerstner2014neuronal} neuron model is the most commonly used neuron in SNNs due to its simplicity. In recent years, several improved LIF-based neurons have been proposed to better adapt to modern artificial intelligence tasks.  
The PLIF neuron \citep{fang2021incorporating} introduces learnable membrane time constants to capture dynamic neuronal behavior. 
PSN \citep{fang2023parallel} further extends LIF by removing the reset mechanism and incorporating a learnable parameter matrix, generating hidden states that are independent across time steps and enabling parallelizable neuronal dynamics. 
LIF, PLIF, and PSN represent key approaches within the spike training and spike inference paradigm.  
An alternative training approach is the integer training and spike inference paradigm. 
ILIF \citep{luo2024integer} implements this paradigm by employing integer-valued activations during training while maintaining spike-driven inference through virtual timesteps. 
NILIF \citep{lei2025spike2former} introduces a normalized version of ILIF to improve training stability for SNNs with complex architectures. 
SpikingBrain \citep{pan2025spikingbrain} further adopts an adaptive threshold in ILIF by precomputing the average membrane potential based on the input; however, this input-dependent adaptation is not well-suited to the spike-driven nature of neuromorphic computation.
Previous works on neurons primarily validated them in a single modality. In this work, we will simultaneously validate the effectiveness and generalizability of our methods in both visual and language modalities.

% \begin{figure}[tb]
% \centering
% \includegraphics[width=0.48\textwidth]{figs/mem_2.pdf}
% \caption{The shiftable characteristics for the membrane potential in the neuron group. (a) shows ILIF. (b) shows the shiftable characteristics can adaptively deliver more information.}
% \label{figure2}
% \end{figure}

\section{Method}
\subsection{A Functional View of Spiking Neurons}
We start with a review of the principles of the three most representative neurons: LIF, ILIF, and PSN.

\textbf{LIF Model.} The dynamics of LIF can be formulated as follows:
\begin{align}
& U[t]=H[t-\mathbf{1}]+X[t], \\
& S[t]=\Theta\left(U[t]-V_{t h}\right), \\
& H[t]=\beta(U[t]-S[t]),
\end{align}
where $t$ denotes the timestep, and $X[t]$ is the input current at time step $t$. When the membrane potential $U[t]$ exceeds the firing threshold $V_{th}$, the spiking neuron will trigger a spike $S[t]$. $\Theta(v)$ is the Heavisine step function, which equals to 1 when $v\geq 0$ and 0 otherwise. 
$U(t)$ will subtract the spike and then decays to $H(t)$ by a factor of $\beta$.  Shown in Figure \ref{main_fig} (d), LIF employs spike training and spike inference paradigm, in which neuronal activity is constrained within a fixed range of [0, 1].  
Owing to its T-dimensional temporal expansion, LIF neuron \cite{gerstner2014neuronal}  suffers from low training efficiency during training, making it more suitable for small to medium-scale SNN models. Moreover, LIF lacks both the capability for efficient training and adaptive dynamics.

\textbf{ILIF Model.} The dynamics of ILIF are formulated as follows:
\begin{align}
&U[t]=H[t-1]+X[t], \notag \\
&S[t]=\operatorname{c l i p}(\operatorname{round}(U[t]), 0, D) , \\
&H[t]=\beta(U[t]-S[t] ), \notag
\end{align}
where $\operatorname{clip}(U[t], min, max)$ denotes  the operation of clipping $U[t]$ to $[min, max]$, $D$ indicates the maximum quantized integer value and unfold into $D$ time steps when inference on neuromorphic chips. That is, the total time step $\operatorname{T}$ of the spike sequence is $T*D$, where $T$ is the normalized integer time step. For example, ILIF(1$\times$4) unfolds into a binary spike sequence of time step $\operatorname{T}$ = 4.
Shown in Figure \ref{main_fig} (e),  ILIF and its variant NILIF\citep{lei2025spike2former} employ an integer-training and spike-inference way \cite{luo2024integer},  where the membrane potential is clamped to a fixed interval of [0, D] when training. Although both ILIF and its variants enable efficient training, the absence of adaptive neuronal dynamics limits their potential as universal spiking neuron models.

\textbf{PSN Model} \cite{fang2023parallel} eliminates the reset process in LIF and achieves adaptive dynamics by applying a linear transformation to the T-dimensional input membrane potential using a $T^2$ parameter matrix.
\begin{align}
&\boldsymbol{H}  =\boldsymbol{W} \boldsymbol{X}, & & \boldsymbol{W} \in \mathbb{R}^{T \times T}, \boldsymbol{X} \in \mathbb{R}^{T \times N}, \\
&\boldsymbol{S}  =\Theta(\boldsymbol{H}-\boldsymbol{B}), & & \boldsymbol{B} \in \mathbb{R}^T, \boldsymbol{S} \in\{0,1\}^{T \times N}, 
\end{align}
where $\boldsymbol{X}$ is the input sequence, $\boldsymbol{w}$ is the learnable weight matrix, $\boldsymbol{H}$ is the hidden state sequence, $\boldsymbol{B}$ is the learnable threshold, and $\boldsymbol{S}$ is the binary output spike sequence.  While PSN retains the T-dimensional training simulation, resulting in training efficiency that lies between that of LIF and ILIF. More critically, the parameter matrix of PSN is difficult to optimize, making its integration with Spiking Transformers an unresolved challenge and thereby limiting its general applicability.

Based on the above analysis, we present a novel functional perspective on the new generation of spiking neuron design to better accommodate vision and language modeling. 
% We argue that the four basic characteristics of spiking neurons must be considered simultaneously.  We argue that a universal spiking neuron should exhibit the following key properties:
We argue that a next-generation foundational spiking neuron should simultaneously satisfy four essential characteristics:
% Existing spiking neuron models struggle to simultaneously achieve both efficient training and adaptive capability, thereby limiting their versatility in SNN training. We argue that a universal spiking neuron should exhibit the following key properties:

1) \textbf{Efficient Training}: The neuron should support efficient SNN training, making it suitable for large-scale models and modern research settings.

2) \textbf{Adaptive Firing}: The neuron should incorporate adaptive Firing with respect to the membrane potential to avoid excessive or insufficient spike firing.
%Membrane Potential Adaptability

3) \textbf{Architecture Compatibility}: The neuron should demonstrate cross-domain generalization (e.g., vision and language) while maintaining compatibility with advanced architectures such as spiking transformers.

4) \textbf{Spike-Driven Inference}: The neuron should conform to a spike-driven computational paradigm when inference to ensure high energy efficiency in SNNs.

We present a comparison of mainstream spiking neurons in Table \ref{neuron_complexity}. Our ASN family (including ASN and NASN) satisfies the four characteristics.

% Based on the above analysis, we provide a Novel Functional View For universal Spiking Neuron to adapt to current artificial intelligence tasks.
% Previous spiking neurons could not simultaneously possess both speed and adaptability, which limited their versatility in SNN training.
% Modern artificial intelligence is characterized by multi-tasking and large-scale parameters. Traditional spiking neurons struggle to handle these tasks.
% We advocate that a universal spiking neuron should possess the following four characteristics:

% 1) \textbf{Rapid Spike Simulation}: Fast spiking simulation is suitable for scientific research in the era of large-scale models.

% 2) \textbf{Adaptive}: The neuron should possess certain adaptive properties to the membrane potential.

% 3) \textbf{Spike-Driven}: The neuron need satisfy the characteristics of spike-driven.

% 4) \textbf{Generality}: The neuron need to have cross-task versatility.

%ILIF \cite{luo2024integer} and its extended neuron models \cite{lei2025spike2former, pan2025spikingbrain} can quickly simulate spiking, but the lack of neuronal adaptation limits their potential to become universal spiking neurons.
%
%PLIF \cite{fang2021incorporating} performs an adaptive transformation of the membrane potential using the sigmoid function; however, the sigmoid function increases training costs, especially with the T-dimensional expansion. Therefore, PLIF's training cost is even higher than LIF's.
%

\subsection{Adaptive Spiking Neuron}
We propose the Adaptive Spiking Neuron (ASN), a general spike-driven neuron model that simultaneously achieves efficient training and adaptive membrane potential dynamics.
By adopting the integer training and spike inference paradigm in ILIF, our ASN emits integer values during training and unfolds them into 0/1 spikes during inference, leading to the extreme efficient training. 
On the other hand,  ASN introduces a layer-wise trainable parameter  $\alpha$ to adaptively shift the activation window, enabling better alignment with the data distribution across different layers. Shown in Figure \ref{main_fig} (c) and (f), the learnable offset facilitates adaptive spike firing regulated by the membrane potential. Accordingly, the proposed ASN can be formulated as follows:
\begin{align}
&U[t]=H[t-\mathbf{1}]+X[t], \notag \\
&S[t]=\operatorname{clip}(\operatorname{round}(U[t]), \alpha, \alpha+D), \label{ASN}   \\
&H[t]=\beta(U[t]-S[t]), \notag
\end{align}
% \textbf{Integer-Valued Training.} 
where $\operatorname{round}(*)$ rounds to the nearest integer, $\operatorname{clip}(X, min, max)$ denotes that clipping $X$ to $[min, max]$. $\alpha$ is the learnable shift parameter. Since the rounding and clipping operations in Eq.\ref{ASN} is discontinuous, the gradient optimization of the input $x$ and $\alpha$ is shown in Section \ref{gradient}.

% Since Eq.\ref{ASN} is discontinuous, its derivative becomes a step function, which may lead to training instability. 
%To address this issue, prior studies have proposed various surrogate gradient functions, primarily designed for binary spike outputs. 
% In our implementation, we consistently adopt a rectangular window as the surrogate function.
% We retain gradients solely for neurons activated in the $[\alpha, \alpha+D]$ range, nullifying all others. 

%In addition, we use the mean of the global gradient to update the trainable parameter $\alpha$.

% \textbf{Spike-Driven Inference.} 

% \begin{equation}
% \begin{aligned}
% \mathbf{U}^{l+1}&=\mathbf{W}^{l+1} \operatorname{clip}(\operatorname{round}(\mathbf{U}^{l}), \alpha, \alpha+D) \\
% &=\mathbf{W}^{l+1} \operatorname{clip}(\operatorname{round}(\mathbf{U}^{l}), 0, D) + \mathbf{W}^{l+1} \lceil \alpha  \rceil
% \end{aligned}
% \end{equation}
The synaptic computation within a layer is illustrated as 
\begin{equation}
\mathbf{X}^{l+1}=\mathbf{W}^{l}S^l[t]=\mathbf{W}^{l} \operatorname{clip}(\operatorname{round}(\mathbf{U}^{l}), \alpha, \alpha+D),
\end{equation}
where $\mathbf{X}^{l}$ means the membrane potential before quantization at $l$-th layer, $\mathbf{W}^{l}$ denotes the synaptic weights in $(l)$-th layer. Similar to ILIF, we convert the integer values to binary spikes in the inference stage. Linear computation in inference can be divided into spike-based computation and the addition of a constant term, which is shown as follows:
\begin{equation}
\mathbf{X}^{l+1}=\mathbf{W}^{l} {S_{1}^{l}[t] + \operatorname{C}}.
\end{equation}
We extend the T time step to $T\times D$, and convert the integer value $S_{1}^l[t]$ to a spike sequence $\left\{S^l[t, d]\right\}_{d=1}^D$, which satisfied:
\begin{equation}
\sum_{d=1}^D S^l[t, d]=S_{1}^l[t].
\label{s(t)}
\end{equation}
The constant term can be formulated as
\begin{equation}
\mathbf{W}^{l} \lceil \alpha  \rceil = \operatorname{C},
\end{equation}
where $\lceil *\rceil$ means round up operation. Compared to the inference process of ILIF, ASN only needs to add a constant $\operatorname{C}= \mathbf{W}^{l+1} \lceil \alpha  \rceil$ to each layer.

\subsection{Normalized Adaptive Spiking Neuron}
Similar to NILIF \citep{lei2025spike2former}, ASN may also suffer from gradient instability due to integer values.  We introduce a more versatile variant by approximately normalizing the virtual step size $D$, which stabilizes both numerical values and gradients. This variant is termed the Normalized Adaptive Spiking Neuron (NASN). Its dynamic behavior is described as follows:
\begin{align}
&U[t]=H[t-\mathbf{1}]+X[t], \notag \\
&S[t]=\operatorname{clip}(\operatorname{round}(U[t]), \alpha, \alpha+D) / N, \\
&H[t]=\beta\left(U[t]-S[t]^* N\right).
\end{align}
For simplicity, we usually set $N=D$. The synaptic computation within a layer is illustrated as follows:
% \begin{equation}
% \begin{aligned}
% \mathbf{U}^{l+1} 
% &= \mathbf{W}^{l+1} (\frac{\mathbf{1}}{N} \operatorname{clip}(\operatorname{round}(\mathbf{U}^{l}), \alpha, \alpha+D)) \\
% &=\frac{\mathbf{1}}{N} \mathbf{W}^{l+1} \operatorname{clip}(\operatorname{round}(\mathbf{U}^{l}), 0, D)\\  & \quad  + \frac{\mathbf{1}}{N}\mathbf{W}^{l+1} \lceil \alpha  \rceil
% \end{aligned}
% \end{equation}
\begin{equation}
\mathbf{X}^{l+1} =\mathbf{W}^{l}S^l[t]= \mathbf{W}^{l} (\frac{\mathbf{1}}{N} \operatorname{clip}(\operatorname{round}(\mathbf{U}^{l}), \alpha, \alpha+D)). \\
\end{equation}
Similar to ASN, the linear computation in Normalized ASN can also be divided into spike-based computation and the addition of a constant term in the inference stage, which is shown as
\begin{equation}
\mathbf{X}^{l+1}=\mathbf{W_{1}}^{l} {S_{1}^{l}[t] + \operatorname{C}},
\end{equation}
where ${S_{1}^{l}[t]}$ is same as Eq. \ref{s(t)}. The weight needs to be scaled as follows:
\begin{equation}
\mathbf{W_{1}}^{l}=\frac{\mathbf{1}}{N}\mathbf{W_{1}}^{l}.
\end{equation}
The constant term can be formulated as follows:
\begin{equation}
\operatorname{C}=\frac{\mathbf{1}}{N}\mathbf{W}^{l+1} \lceil \alpha  \rceil.
\end{equation}
Compared to the inference process of NILIF, NASN only needs to add a constant $\operatorname{C}$ to each layer.

% Unlike conventional spiking neuron that employ fixed-interval clamping, our ASN family (including ASN and NASN) featuring a learnable offset parameter $\alpha$, can dynamically shift its quantization window to better align with the evolving distribution of membrane potentials across different layers, enhancing the representational capacity and distributional adaptability of Spiking Neural Networks (SNNs)

Unlike conventional spiking neurons that employ fixed-interval clamping, our ASN family (including ASN and NASN) incorporates a learnable offset parameter 
$\alpha$, enabling dynamic shifts of the quantization window to better align with the evolving distribution of membrane potentials across layers. This mechanism enhances both the representational capacity and distributional adaptability of SNNs.

\begin{table*}[!tb]
  \centering
  % \vspace{+0.2cm}
    \caption{The results on ImageNet-1k classification. "E (mJ)" means "Energy Consumption (mJ)". Spikingformer$^\dagger$ is the variant of Spikingformer with the improved downsampling \cite{zhou2023enhancing}.}
    \begin{tabular}{llccp{1.5cm}<{\centering}p{1.5cm}<{\centering}p{1.3cm}<{\centering}p{1.8cm}<{\centering}}
    \toprule
    \multicolumn{1}{l}{Methods} & \multicolumn{1}{l}{Architecture} & Param{} (M)&Train Size &Test Size & {Time Step} & E (mJ)  & {Top-1 Acc ($\%$)} \\
    \midrule
    SEW ResNet\cite{fang2021deep}  & SEW-ResNet-34 &21.79 &224$^2$ &224$^2$  &4 &-  & 67.04 \\
    {MS-ResNet\cite{hu2021advancing}}  &{MS-ResNet-34} &21.80 &224$^2$ &224$^2$ & 6  &-  & {69.42}\\
   Spikformer\cite{zhou2023spikformer}   &{Spikformer-8-384} &16.81 &224$^2$ &224$^2$ &4 &{12.43}  & 70.24\\
    SD-Transformer\cite{yao2023spike}  & {SD-Transformer-8-384} &16.81 &224$^2$ &224$^2$  &4 &3.90  & {72.28}\\
    LIF (Spikingformer)\cite{zhou2026spikingformer} &{Spikingformer-8-384} &16.81 &224$^2$ &224$^2$ &4 &{4.69}  & {72.45}\\
    LIF (Spikingformer$^\dagger$)\cite{zhou2026spikingformer} &{Spikingformer-8-384} &16.81 &224$^2$ &224$^2$ &4 &{5.61}  & {74.35}\\
    NILIF (Spikingformer$^\dagger$) &{Spikingformer-8-384} &16.81 &224$^2$ &224$^2$ &4 &{5.58}  & {75.41}\\
    \midrule
    % {\multirow{3}{*}{\textbf{Ours (A-LIF)}}}
    % &{Spikingformer-8-384} &16.81 &224$^2$ &224$^2$ &4 &{5.61}  & {todo}\\
    % &{Spikingformer-8-512}&29.68 &224$^2$ &224$^2$ & 4 &{8.68}  & {todo}\\
    % &{Spikingformer-8-768}&66.34 &224$^2$ &224$^2$ & 4 &{16.30}  & {todo}\\
    % \cmidrule(lr){2-8}
    {\multirow{1}{*}{\textbf{Ours}}}
    &{Spikingformer-8-384} &16.81 &224$^2$ &224$^2$ &4 &{5.67}  & \textbf{75.53}\\
    \bottomrule
    \end{tabular}%
    % 
  % \vspace{-4mm}
  \label{tab:imagenet}%
\end{table*}%

\begin{table}[!tb]
  \centering
  %$^\dagger$ means Spikingformer in CML way \cite{zhou2023enhancing}.
  % \vspace{+0.2cm}
    \caption{Comparision on CIFAR-10 and CIFAR-100. "Param" denotes "Parameter (M)", "Acc" denotes "Top-1 Accuracy (\%)", and "$T$" denotes "Time Step". Our model adopts the architecture of Spikingformer$^\dagger$. } 
    \begin{tabular}{lp{0.8cm}<{\centering}p{0.1cm}<{\centering}p{0.8cm}<{\centering}p{0.8cm}<{\centering}p{0.1cm}<{\centering}p{0.8cm}<{\centering}}
    \toprule
    \multicolumn{1}{c}{\multirow{2}[4]{*}{Method}} & \multicolumn{3}{c}{CIFAR-10} & \multicolumn{3}{c}{CIFAR-100}  \\
\cmidrule(l{2pt}r{2pt}){2-4}\cmidrule(l{2pt}r{2pt}){5-7}
    &{Param} & {$T$} & {Acc} &Param & {$T$} & {Acc} \\
    \midrule
    % Hybrid training &9.27 &125 &92.22 &9.27 &125 &67.87 & $-$ & $-$ & $-$ & $-$ & $-$ & $-$\\
    % TET &12.63 & 4 & 94.44 &12.63 & 4 & {74.47} & $-$ & $-$ & $-$ & $-$ & $-$ & $-$\\
    % PLIF  & $-$ & $-$ & $-$ & $-$ & $-$ & $-$  & $-$ & 20 & {74.8} & $-$ & {20} & {97.6} \\
    % SEW ResNet\cite{fang2021deep}  & $-$ & $-$ & $-$ & $-$ & $-$ & $-$ \\
    % STBP-tdBN  &12.63 & 4 & 92.92 &12.63 & 4  & 70.86 & $-$ & $-$  & 96.87 & $-$ & $-$  & 67.8 \\
    % MS-ResNet  &$-$ & $-$ & 91.72 & $-$  & $-$ & $66.83$   \\
    Spikformer\cite{zhou2023spikformer}   &5.76 & 4 & 94.80 & 5.76  & 4    & 76.95   \\
    Spikformer\cite{zhou2023spikformer}   &9.32 & 4 & 95.51 & 9.32  & 4    & 78.21   \\
    %\midrule
    SD-Transformer\cite{yao2023spike}  &   10.28    & 4  &  95.60  &   10.28    &   4    &  78.4  \\
    %\midrule
    % STSA  &  $-$  &$-$ & $-$   &  $-$    &  $-$   & $-$      \\
    % QKFormer\cite{zhou2024qkformer}  &   6.74    & 4  &  {96.18}  &   6.74    &   4    &  {81.15}  \\
    % \cmidrule(lr){2-13}
    % ResNet-19 (ANN)  &  12.63  &1 & 94.97   &  12.63    &  1   & 75.35   &    $-$   &   $-$    &  $-$     &   $-$    &  $-$     & $-$ \\
    % \midrule
    Spikingformer\cite{zhou2026spikingformer} &   9.32    & 4  &  95.81  &   9.32    &   4    &  79.21  \\
    Spikingformer$^\dagger$\cite{zhou2026spikingformer} &   9.32    & 4  &  {95.95}  &   9.32    &   4    &  {80.37}    \\
    Transformer(ANN)  &  9.32  &1 & 96.73   &  9.32    &  1   & 81.02   \\
    % \cmidrule(lr){2-13}
    \midrule
    \textbf{Ours}
    &   9.32    & 4  &  \textbf{96.27}  &   9.32    &   4    &  \textbf{81.11}  \\
    \bottomrule
    \end{tabular}%
    % \vspace{-2mm}
  \label{tab:small_dataset}%
\end{table}%

\subsection{Gradient Optimization} \label{gradient}

% \subsection{Adaptive Multi-Spike Neuron (ASN) with Learnable Offset}

\paragraph{Gradient Estimation via STE}
Since the rounding and clipping operations are non-differentiable (possessing zero derivatives almost everywhere), we employ the {Straight-Through Estimator (STE)} to facilitate end-to-end backpropagation. 

\paragraph{Gradient of the Input $x$} 
The gradient is permitted to flow only within the active quantization range. We denate the lower bound $d_{\min}=\alpha$ and the upper bound $d_{\max}=\alpha + D$ . Then, the surrogate gradient for $x$ is defined as:
\begin{equation}
    \frac{\partial y}{\partial x} = 
    \begin{cases} 
    1, & \text{if } d_{\min} <= x <= d_{\max}, \\
    0, & \text{otherwise}.
    \end{cases}
\end{equation}

\paragraph{Gradient of the  Learnable Parameter $\alpha$} 
The optimization of $\alpha$ is primarily driven by samples falling into the truncated region (outside the boundaries). When $x$ resides at or beyond the boundaries, the output $y$ becomes a direct function of the boundary values, yielding the following approximation:
\begin{equation}
    \frac{\partial y}{\partial \alpha} = 
    \begin{cases} 
    1, & \text{if } x < d_{\min} \text{ or } x > d_{\max}, \\
    0, & \text{otherwise}.
    \end{cases}
\end{equation}
Consequently, the total gradient of the objective loss $\mathcal{L}$ with respect to $\alpha$ is accumulated over the mini-batch as:
\begin{equation}
    \nabla_{\alpha} \mathcal{L} = a * \sum_{i} \frac{\partial \mathcal{L}}{\partial y_i} \cdot \mathbb{I}(x_i \notin [d_{\min}, d_{\max}]),
\end{equation}
where $\mathbb{I}(\cdot)$ is the indicator function, and $a$ is a hyperparameter for adjusting gradient scaling. This gradient mechanism allows the network to adaptively shift its activation window to minimize task-specific loss, providing a flexible balance between quantization error and dynamic range preservation. 

%In our main experiments, we mainly use NASN by default.
To facilitate experimental evaluation, NASN ($1\times4$) is used as the default configuration throughout in our main experiments. Note that NASN with the setting of $1\times4$ is equal to 4 time steps.

\begin{table*}[!tb]
\begin{center}
\begin{small}
\centering
\caption{The results on the  Natural Language Understanding task (GLUE datasets). "Avg." denotes "Average Accuracy (\%)". 
The results of LIF-BERT and PSN-BERT are reported in \citet{xing2024spikelm}. Our model adopts the architecture of WE-SpikingFormer.
}
\label{tab_glue}
% \vspace{8pt}
% \setlength{\tabcolsep}{1.7mm}
\fontsize{9.3pt}{\baselineskip}\selectfont
\begin{tabular}{lp{1.6cm}<{\centering}|cccccccc|p{2.0cm}<{\centering}}
\toprule
{Model}  & {Time Step}  & {MNLI} & {QQP} & {QNLI} & {SST-2} & {CoLA} & {STS-B} & {MRPC} & {RTE} & Avg. \\ 
\midrule
BERT$_\texttt{base}$ \citep{devlin2019bert}   & --   & 83.4 & 71.2 & 90.5 & 93.5 & 52.1 & 85.8 & 88.9 & 66.4 & 79.6 \\
Q2BERT \citep{zhang2020ternarybert}  & --   & 47.3 & 67.0 & 61.3 & 80.6 & 0.0  & 4.7  & 81.2 & 52.7 & 49.1 \\
% ELMo                       & --   & 68.6   & 86.2 & 71.1 & 91.5 & 44.1 & 70.4 & 76.6 & 53.4 & 70.2 \\
BERT$_\texttt{3L}$ \citep{devlin2019bert}    & --   & 77.1 & 85.2 & 85.8 & 88.1 & 31.7 & 85.7 & 86.4 & 66.4 & 75.9 \\
SpikeLM \citep{xing2024spikelm}  & 4   & 77.2 & 83.9 & 85.3 & 87.0 & 38.8 &  84.9 &  85.7 & 69.0 & 76.5 \\
\midrule
% LIF-BERT   & 4   & 55.2   & 70.0 & 60.6 & 80.6 & 14.6 & 20.0 & 82.3 & 53.8 & 54.9 \\
% SpikingFormer$^\dagger$   & 4   & 70.6   & 80.9 & 79.5 & 83.9 & 12.8 & 77.0 & 83.0 & 62.1 & 68.9 \\
LIF-BERT \citep{xing2024spikelm}   & 4   & 35.2   & 0 & 50.5 & 50.9 & 0 & 0 & 81.2 & 52.7 & 34.6 \\
PSN-BERT \citep{xing2024spikelm}   & 4   & 35.2   & 0 & 50.5 & 50.9 & 0 & 6.8 & 81.2 & 52.7 & 34.7 \\
SpikeBERT \citep{lv2023spikebert}   & 4    & 71.0 & 68.2 & 66.4 & 85.4 & 16.9 & 18.7 & 82.0 & 57.5 & 59.7 \\
% SpikingFormer          & 4    & 67.6 & 70.1 & 68.9 & 80.3 & 8.9 & 16.7 & 79.4 & 51.3 & 55.4 \\
% \midrule
% SpikingFormer\cite{zhou2026spikingformer}       & 4    & 72.5 & 84.7 & 76.0 & 87.2 & 24.4 & 54.5 & 79.7 & 55.6 & 66.8 \\
NILIF (WE-SpikingFormer) \citep{zhou2026winnertakeallspikingtransformerlanguage}   &  4    & 70.1 & 85.1 & 77.5 & 89.0 & 27.9 & 42.8 & 81.6 & 55.9 & 66.3 \\
\midrule
\textbf{Ours}    & 4  & 75.5 & 86.1 & 82.9 & 87.0 & 33.5 & 45.0 & 79.1 & 50.9 & \textbf{67.5} \\
\bottomrule
\end{tabular}
\end{small}
\end{center}
\end{table*}

\section{Experiments}
% To evaluate the effectiveness of the proposed neuron, we conduct experiments on a diverse set of visual and language modeling tasks across 19 datasets. These benchmarks include the large-scale vision dataset ImageNet~\cite{deng2009imagenet}, the medium-scale vision datasets CIFAR~\cite{krizhevsky2009learning} (including CIFAR-10 and CIFAR-100), as well as a variety of natural language understanding, question answering, and commonsense reasoning tasks.
To validate the effectiveness and generalization of the proposed adaptive neuron, we conduct extensive experiments across a diverse suite of 19 datasets covering both vision and language modeling. Specifically, our evaluation encompasses large-scale image classification on ImageNet-1K \cite{deng2009imagenet}, as well as medium-scale analysis on CIFAR-10/100 \cite{krizhevsky2009learning}. To further demonstrate its versatility, we extend our evaluation to a wide array of linguistic tasks, including Natural Language Understanding, Question Answering, and Commonsense Reasoning, providing a holistic assessment of the neuron's performance across different modalities and scales.

\subsection{ImageNet-1k Classification}\label{sec:imagenet}
In this part, we conduct experiments on the ImageNet-1K dataset, which contains approximately 1.28 million training images and 50,000 validation images across 1,000 object classes. 
We benchmark our model against several state-of-the-art SNN foundantion architectures, including CNN-based foundations such as {SEW ResNet} \cite{fang2021deep} and {MS-ResNet} \cite{hu2021advancing}, as well as Transformer-based SNN models like {Spikformer} \cite{zhou2023spikformer}, Spike-driven Transformer (SD-Transformer) \cite{yao2023spike}, and {Spikingformer} \cite{zhou2026spikingformer}. 
%The results on ImageNet-1K is shown in Tab. \ref{tab:imagenet}.
%We compared our model with SEW ResNet \cite{fang2021deep}, MS-ResNet\cite{hu2021advancing}, Spikformer \cite{zhou2023spikformer}, Spike-Driven Transformer (SD-Transformer) \cite{yao2023spike}, and Spikingformer\cite{zhou2026spikingformer} on ImageNet-1k, which is shown in Tab. \ref{tab:imagenet}.

Among these baselines, Spikingformer represents a robust backbone that integrates spike-driven operations with global modeling capabilities, currently achieving leading performance on ImageNet-1K. For a rigorous and fair comparison, we adopt the {Spikingformer-8-384} architecture as our structural framework, replacing the standard spiking neurons with our proposed NASN. As summarized in Tab.~\ref{tab:imagenet}, 
Our model achieves 75.53$\%$ Top-1 accuracy, exceeding the baseline LIF-based Spikingformer$^\dagger$ by 1.18$\%$ and outperforming NILIF-based Spikingformer$^\dagger$ by 0.12$\%$ under the same architectural settings.
These results underscore the efficacy of the learnable offset $\alpha$ in optimizing the firing threshold to better capture the complex data distributions of large-scale visual tasks.

% SEW ResNet and MS-ResNet are CNN-based SNN foundation models. Spikformer, SD-Transformer, and Spikingformer are Transformer-based SNN foundation models that can be applied to both vision and language tasks.  Among these foundation model, Spikingformer is the backbone that contains both spike-driven features and global modeling capabilities, achieving the best performance in ImageNet-1K. Then, our ASN adopts the architecture of Spikingformer-8-384, and achieves 75.53$\%$ top-1 accuracy on ImageNet-1k. Our model outperforms Spikingformer$^\dagger$ by 1.18$\%$ under the same architecture.

\subsection{CIFAR Datasets}
% We also evaluate our method on the CIFAR-10 and CIFAR-100 datasets, which consist of 60,000 color images of size 32 $\times$ 32. Both datasets share the same training and testing split of 50,000 and 10,000 images, respectively, but differ in complexity with 10 and 100 object categories. These benchmarks are widely used to evaluate model performance on medium-scale image classification tasks.

% We compared our method with  SEW ResNet \cite{fang2021deep}, Spikformer \cite{zhou2023spikformer}, SD-Transformer \cite{yao2023spike} and Spikingformer\cite{zhou2026spikingformer}. The results are shown in Tab. \ref{tab:small_dataset}. 
% In CIFAR-10, our ASN adopts the same architecture and experimental setting as Spikingformer$^\dagger$. Our method achieves 96.27$\%$ accuracy and significantly outperforming Spikingformer$^\dagger$ by 0.32$\%$.
% In CIFAR-100, our ASN adopts the same architecture and experimental setting as Spikingformer$^\dagger$. Our method achieves 81.11$\%$ accuracy and significantly outperforming Spikingformer$^\dagger$ by 0.74$\%$. 

% Transformer (ANN) is only 0.26$\%$ and 0.02$\%$ higher than Spikingformer$^\dagger$ on CIFAR-10 and CIFAR-100. 

To further validate the effectiveness and generalization of our method, we conduct experiments on the {CIFAR-10} and {CIFAR-100} datasets \cite{krizhevsky2009learning}. Both datasets consist of 60,000 RGB images with a resolution of $32 \times 32$, partitioned into 50,000 training and 10,000 testing samples. While they share the same data scale, CIFAR-100 presents a higher degree of classification complexity with 100 object categories compared to the 10 categories in CIFAR-10. These benchmarks serve as standard metrics for evaluating model performance on medium-scale visual recognition tasks.

We compared our method with Spikformer \cite{zhou2023spikformer}, SD-Transformer \cite{yao2023spike} and Spikingformer\cite{zhou2026spikingformer}. The quantitative results are detailed in Tab.~\ref{tab:small_dataset}. For a consistent evaluation, we adopt the identical architecture and experimental configurations as the baseline {Spikingformer}$^\dagger$. 
On CIFAR-10, our model achieves an accuracy of {96.27\%}, surpassing the Spikingformer$^\dagger$ baseline by {0.32\%}. On the more challenging CIFAR-100 dataset, our method achieves {81.11\%} accuracy, yielding a more pronounced improvement of {0.74\%} over Spikingformer$^\dagger$. 

%These consistent gains across datasets of varying complexity demonstrate that our adaptive offset mechanism effectively mitigates quantization errors and enhances the discriminative feature learning of the spiking backbone.

\begin{table*}[htb]
\begin{center}
\begin{small}
\centering
\setlength{\tabcolsep}{1.7mm}
\fontsize{9.3pt}{\baselineskip}\selectfont
\caption{The results on Question Answering Tasks (QAT). "E (mJ)" means "Energy Consumption (mJ)". "Avg." denotes "Average Accuracy (\%)". 
Our model adopts the architecture of WD-SpikingFormer-0.4B. }
\label{QA_tasks}
% \vspace{8pt}
\begin{tabular}{lp{1.0cm}<{\centering}p{2.0cm}
<{\centering}|ccccc|p{2.0cm}<{\centering}}
\toprule
{Model} & E (mJ) & {Time Step}  & {ARC-e} & {ARC-c} & {BoolQ} & {HeadQA}  & {OBQA} & Avg. \\ 
\midrule
% SpikeLLM-7B \citep{xing2024spikellm} & -  & 4 & 31.3 & 23.6   & 53.8 & -  & -  & -  \\
Qwen-1.5B \citep{qwen2} & 3398.3  & - & 26.2 & 26.9   & 60.3 & 25.7  & 27.1 & 33.2  \\

% \textbf{WD-SpikingFormer-0.4B$\_\operatorname{sm}$}  & -    & 1$\times$4    &30.4  & 21.6  & 37.8 & 27.1 & 26.6 & \textbf{28.7} \\
NILIF (WD-SpikingFormer-0.4B) \citep{zhou2026winnertakeallspikingtransformerlanguage}   & 238.4    & 4    &30.0  & 22.3 & 37.8 & 26.0 & 26.1 & {28.4} \\
% {SpikingFormer\_NILIF}       & 238.4    & 4    &30.0  & 22.3 & 37.8 & 26.0 & 26.1 & \textbf{todo} \\
% {WD\_SpikingFormer} \citep{zhou2026winnertakeallspikingtransformerlanguage}  & 238.4    & 4    &30.0  & 22.3 & 37.8 & 26.0 & 26.1 & \textbf{todo} \\
\midrule
\textbf{Ours}         & 245.1    & 4    &35.3  & 20.6 & 37.9 & 25.9 & 27.2 & \textbf{29.4} \\
\bottomrule
\end{tabular}
\end{small}
\end{center}
\end{table*}

\begin{table*}[htb]
% \begin{center}
\begin{small}
\centering
\setlength{\tabcolsep}{1.7mm}
\caption{The results on Commonsense Reasoning Tasks (CRT). "E (mJ)" means "Energy Consumption (mJ)". "Avg." denotes "Average Accuracy (\%)" . 
Our model adopts the architecture of WD-SpikingFormer-0.4B. }
% \vspace{8pt}
\label{commonsense_reasoning_tasks}
\fontsize{9.3pt}{\baselineskip}\selectfont
\begin{tabular}{lp{1.5cm}<{\centering}p{2.0cm}<{\centering}|p{1.5cm}<{\centering}p{1.5cm}<{\centering}p{1.5cm}<{\centering}|p{2.3cm}<{\centering}}
\toprule
{Model} & E (mJ) & {Time Step}  & {HellaSwag} & {PIQA} & {WinoGrande} & Avg. \\ 
\midrule
% LLaMa-7B & -  & - & 73.0 & 77.0 & 67.2  & 72.4   \\
SpikeLLM-7B \citep{xing2024spikellm} & -  & 4 & 33.9 & 53.4 & 51.5  & 46.3   \\
Qwen-1.5B \citep{qwen2} & 3398.3  & - & 26.5 & 53.7 & 52.8  & 44.3  \\
NILIF (WD-SpikingFormer-0.4B) \citep{zhou2026winnertakeallspikingtransformerlanguage}    & 238.4    & 4   & 25.9 & 53.4 & 50.2 & {43.2} \\
\midrule
\textbf{Ours}    & 245.1    & 4   & 26.0 & 54.6 & 49.4 & \textbf{43.3} \\
\bottomrule
\end{tabular}
\end{small}
% \end{center}
\end{table*}

\subsection{Natural language understanding}
We evaluate our method on the GLUE benchmark \cite{wang2018glue}, a widely used suite of natural language understanding tasks designed to assess the general linguistic capabilities of machine learning models across diverse domains.
GLUE consists of eight tasks, and these tasks cover several categories of language understanding, including single-sentence classification (CoLA, SST-2), sentence pair classification and paraphrase detection (MRPC, QQP, RTE), semantic textual similarity (STS-B), and natural language inference (MNLI, QNLI), providing a comprehensive evaluation of a model’s ability to capture syntactic, semantic, and relational information between sentences.
We pretrain the model on the Wikipedia-English corpus \cite{devlin2019bert} using the Masked Language Modeling (MLM). The pretraining stage is conducted on 8 GPUs, following the architectural configurations and experimental settings established by WE-Spikingformer \citep{zhou2026winnertakeallspikingtransformerlanguage}, which uses NILIF neuron. Subsequently, we fine-tune the pretrained model on the GLUE set. 
For a comprehensive comparison, we include Artificial Neural Network (ANN) baselines, specifically the original BERT \cite{devlin2019bert} and the quantized Q2BERT \cite{zhang2020ternarybert}, the latter of which utilizes 2-bit weights and 8-bit activations.
%
% We pretrain our method on Wikipedia-English \cite{devlin2019bert} with masked language modeling \cite{devlin2019bert} using 8 GPUs, and subsequently fine-tune it on the GLUE dev set, which is the same as WE-SpikingFormer \citep{zhou2026winnertakeallspikingtransformerlanguage} on architecture and experimental setting. 
% %
% The ANN baseline includes BERT \cite{devlin2019bert} and Q2BERT \cite{zhang2020ternarybert}, the latter of which employs 2-bit weights and 8-bit activations. 

The experimental results are summarized in Table~\ref{tab_glue}. All evaluated models, with the exception of the 3-layer BERT$_\texttt{3L}$, consist of 12 encoder blocks and maintain a comparable parameter scale of approximately 0.1B. Our method achieves an average accuracy of 67.5\%, outperforming the NILIF version of WE-Spikingformer by 1.2\%. 
In our comparative analysis, we note that while SpikeLM achieves higher performance, it functions as a softmax-based spiking transformer and retains non-spiking GeLU \cite{hendrycks2016gaussian} activations within its MLP blocks. In contrast, our method operates as a softmax-free, spike-driven transformer. Within the category of spike-driven architectures, our method significantly outperforms existing methods such as LIF-BERT (34.6\%), PSN-BERT, and SpikeBERT (59.7\%), achieving a substantial performance gain with an accuracy of 67.5\%. These results demonstrate that the learnable offset $\alpha$ effectively bridges the representational gap in complex language understanding tasks.

% The experimental results are shown in Table~\ref{tab_glue}, and all models in Table~\ref{tab_glue} except BERT$_\texttt{3L}$ have 12 encoder blocks and have the same model size with about 0.1B parameters. 
% %
% Our method achieves $67.5\%$ average accuracy, and outperforms WE-Spikingformer by 1.2$\%$.
% %
% In comparison, SpikeLM is a softmax-based spiking transformer and retains non-spiking activation GeLU \cite{hendrycks2016gaussian} in MLP blocks. These factors led to its relatively higher performance. 
% %
% As a softmax-free spiking transformer, our method outperforms other spike-driven methods (LIF-BERT, PSN-BERT, and SpikeBERT) by a significant margin. ASN vs. SpikeBERT vs. LIF-BERT. Acc: 67.5$\%$ vs. 59.7$\%$ vs. 34.6$\%$.

% \subsection{Zero-shot Learning} \label{Zero-shot Learning}

\subsection{Question Answering Tasks} \label{Question_Answering_Tasks}
We evaluate our model on several Question Answering Tasks (QAT) covering scientific reasoning, Boolean inference, and domain-specific knowledge.
ARC-e and ARC-c \citep{clark2018think} (AI2 Reasoning Challenge, easy and challenge subsets) evaluate scientific knowledge and reasoning abilities. 
ARC-e focuses on relatively straightforward multiple-choice science questions, while ARC-c contains more challenging problems that require advanced reasoning.
BoolQ (Boolean Questions) \citep{clark2019boolq} is a yes/no question-answering dataset derived from naturally occurring queries, where models must determine whether a statement is true or false given a supporting passage.
HeadQA \citep{vilares2019head} is a multilingual medical question-answering benchmark constructed from professional healthcare examinations, designed to assess domain-specific medical knowledge.
OBQA (OpenBookQA) \citep{mihaylov2018can} evaluates a model’s ability to answer elementary science questions by combining a small set of provided core facts with external common knowledge.

We pretrain our model on the FineWeb-Edu corpus \citep{lozhkov2024fineweb-edu}, a high-quality subset of the FineWeb dataset specifically curated for factual and educational content. The experimental configurations follow WD-Spikingformer \citep{zhou2026winnertakeallspikingtransformerlanguage}, and the corresponding results are detailed in Table~\ref{QA_tasks}. Overall, our model achieves an average accuracy of 29.4\% across the evaluated Question-Answering benchmarks, 1.0$\%$ higher than models with an equivalent architecture but utilizing NILIF neurons.
Furthermore, we evaluate the energy efficiency of our approach. Compared to Qwen-1.5B, which utilizes the same pretraining setup as WE-Spikingformer, our model reduces energy consumption by an order of magnitude, requiring only 245.1 mJ compared to 3398.3 mJ (representing only 7\% of the energy). While maintaining this high energy efficiency, our model preserves a narrow accuracy gap (29.4\% vs. 33.2\%) against the 1.5B dense ANN baseline. These results underscore the superior energy-accuracy trade-off of our proposed method.

\subsection{Commonsense Reasoning Tasks}
We evaluate our methods on several Commonsense Reasoning Tasks (CRT) that assess a model’s ability to perform grounded inference, physical reasoning, and context-aware understanding.
HellaSwag \citep{zellers2019hellaswag} is a large-scale benchmark for grounded commonsense inference, where models must select the most plausible continuation of a given context.
PIQA (Physical Interaction Question Answering) \citep{bisk2020piqa} \citep{bisk2020piqa} evaluates physical commonsense reasoning by requiring models to choose the more plausible solution to everyday tasks.
WinoGrande \citep{sakaguchi2021winogrande} tests commonsense reasoning through coreference resolution, requiring models to identify the correct pronoun reference that cannot be resolved by syntax alone.

The experimental configurations for our model are aligned with the protocols established by WD-Spikingformer \citep{zhou2026winnertakeallspikingtransformerlanguage}. The comprehensive results for commonsense reasoning are presented in Table~\ref{commonsense_reasoning_tasks}. Our model achieves an average accuracy of 43.3\% across these tasks, representing a 0.1\% improvement over the baseline WD-Spikingformer with NILIF neuron.
When compared to significantly larger models such as Qwen-1.5B and SpikeLLM-7B, our 0.4B-parameter model demonstrates remarkable parameter efficiency. Despite the substantial disparity in model scale (0.4B vs. 1.5B and 7B parameters), our model maintains a highly competitive performance level, with an average accuracy of 43.2\% compared to 44.3\% for Qwen-1.5B and 46.3\% for SpikeLLM-7B. These findings indicate that the proposed adaptive mechanism allows the model to retain strong reasoning capabilities while utilizing a fraction of the parameters required by dense ANN or larger spiking counterparts.

% The experimental settings of our model are same with WD-SpikingFormer\citep{zhou2026winnertakeallspikingtransformerlanguage}. The experimental results are shown in Table~\ref{commonsense_reasoning_tasks}. Our model achieves 43.3$\%$ average accuracy on commonsense reasoning tasks, outperforming WD-SpikingFormer by 0.1 $\%$.
% %
% Compared with SpikingLLM-7B and Qwen-1.5B, WD-SpikingFormer-0.4B uses smaller parameters (0.4B vs. 1.5B and 7B) while its accuracy remains close to the larger models ($43.2\%$ vs. $44.3\%$ and $46.3\%$). 
% These results highlight that WD-SpikingFormer-0.4B delivers competitive performance despite its much smaller size.

% \subsection{Ablation study} 

\begin{figure}[tb]
\centering
\includegraphics[width=0.48\textwidth]{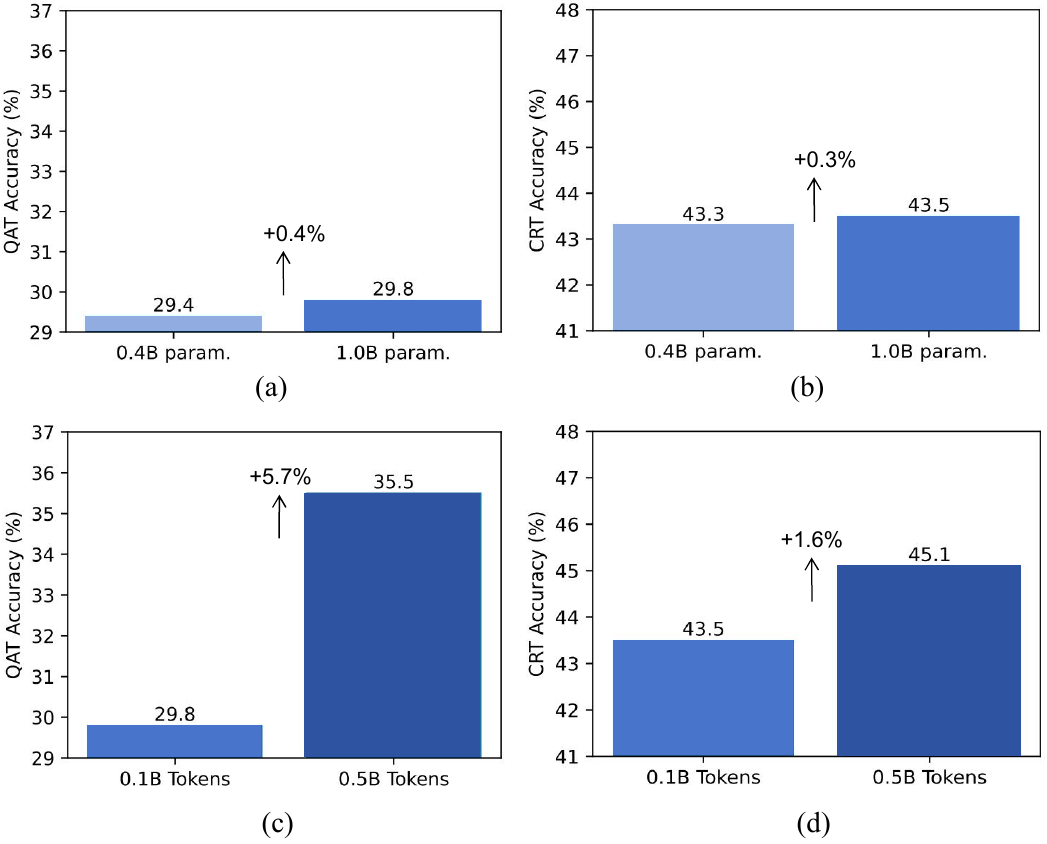}
\caption{Our model (WD-SpikingFormer with NASN) on parameter scaling and pretraining data scaling. (a) and (b) show increasing model parameters for pretraining on Question-Answering Tasks (QAT) and Commonsense Reasoning Tasks (CRT). (c) and (d) show increasing tokens for model pretraining on the above two Tasks. }
\label{scaling}
\end{figure}

\begin{table}[tb]
\begin{small}
\centering
\setlength{\tabcolsep}{1.7mm}
\caption{Ablation study for Spiking Neurons on CIFAR-100. Base model adopts the architecture of Spikingformer$^\dagger$ \citep{zhou2026spikingformer}.  }
\label{ablation_study}
\fontsize{9.3pt}{\baselineskip}\selectfont
\begin{tabular}{p{3.0cm}<{\centering}|p{3.0cm}<{\centering}}
\toprule
Neuron model & CIFAR-100 ($\%$)  \\ 
\midrule
LIF \citep{zhou2026spikingformer}  & 80.22   \\
PLIF \citep{fang2021incorporating} & 80.25   \\
PSN \citep{fang2023parallel} & 77.84   \\
\midrule
ILIF \citep{luo2024integer} & 80.32   \\
ASN (Ours) & 80.36   \\
\midrule
NILIF \citep{lei2025spike2former} & 80.98  \\
NASN (Ours) & 81.11   \\
% \midrule
% Ternary ASN & todo  & todo  \\
\bottomrule
\end{tabular}
\end{small}
\end{table}

\begin{figure}[tb]
\centering
\includegraphics[width=0.48\textwidth]{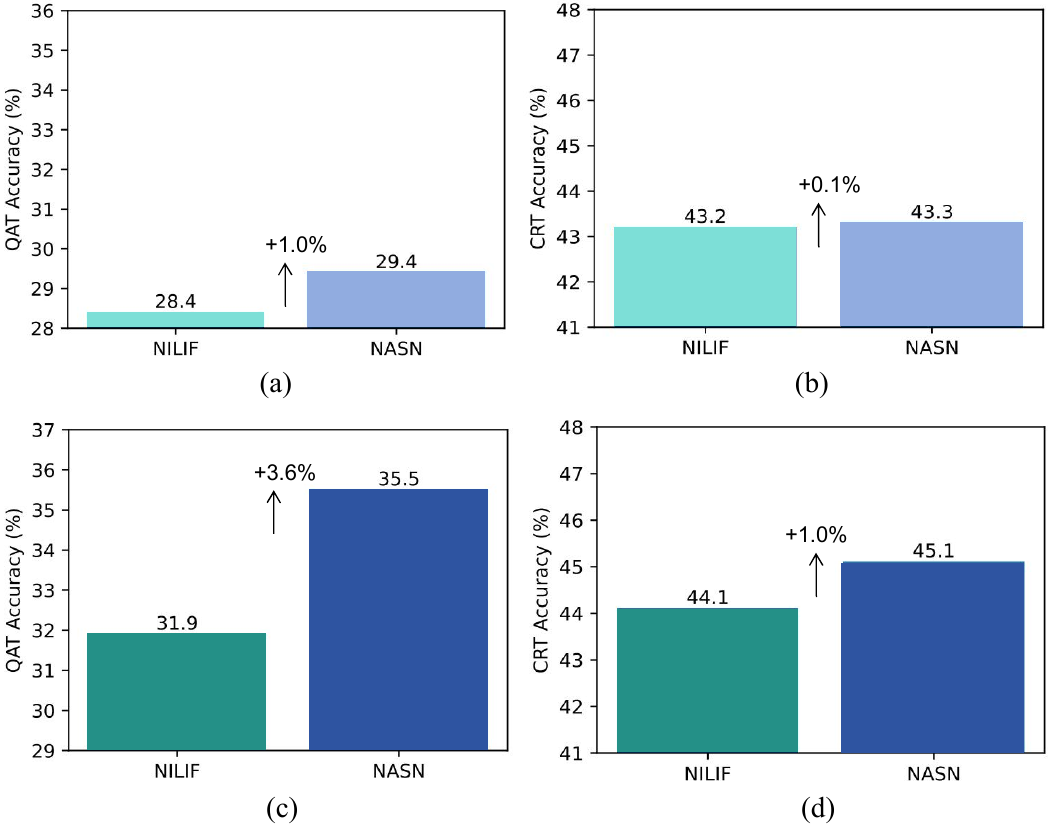}
\caption{Neuron comparison on scaling. The base model is WD-SpikingFormer. (a) and (b) show the neuron comparison with the setting of 0.4B param with 0.1B Tokens on Question-Answering Tasks (QAT) and Commonsense Reasoning Tasks (CRT).
(c) and (d) show the neuron comparison with the setting of 1.0B param with 0.5B Tokens on the above two Tasks. }
\label{Neuron_scaling}
\end{figure}

\section{Discussion and Ablation Study} %model scaling. and limitation.

\subsection{Discussion}
In this part, we discuss the scaling characteristics of our method in the spiking transformers from the model parameter and pretraining data perspectives. The base model is the WD-SpikingFormer architecture with our NASN.

\textbf{Parameter scaling.} The experimental results are shown in Figure \ref{scaling} (a) and (b). We increase the model parameters from 0.4B to 1.0B. The experimental results show that the performance of the model can be further improved on both QAT and CRT. NASN (0.4B) vs. NASN (1.0B). QAT: 29.4$\%$ vs. 29.8$\%$; CRT: 43.3$\%$ vs. 43.5$\%$.

\textbf{Pretraining data scaling.}
The experimental results are presented in Figure \ref{scaling} (c) and (d). In this part, we increase the tokens for NASN (1.0B) pretraining and validate it on QAT and CRT. 
We found that NASN (1.0B) can achieve a further significant improvement on both QAT and CRT. NASN (1.0B) with 0.1B tokens vs. NASN (1.0B) with 0.5B tokens. QAT: 29.8$\%$ vs. 35.5$\%$; CRT: 43.5$\%$ vs. 45.1$\%$.
In particular, by improving the pretraining data of NASN (1.0B) from 0.1B tokens to 0.5B tokens, the performance of NASN (1.0B) achieves a significant performance improvement of 5.7$\%$ on QAT, demonstrating that our methods are capable of adapting to larger-scale pretraining.

% \begin{table}[!tb]
% \begin{small}
% \centering
% \setlength{\tabcolsep}{1.7mm}
% \caption{Ablation study on CIFAR and GLUE datasets. }
% \label{ablation_study}
% \fontsize{9.3pt}{\baselineskip}\selectfont
% \begin{tabular}{lp{3.0cm}<{\centering}p{3.0cm}<{\centering}}
% \toprule
% Dataset & CIFAR-100 ($\%$) & GLUE ($\%$)  \\ 
% \midrule
% ILIF \citep{luo2024integer} & 80.02  & 71.5  \\
% ASN & 80.15  & 73.8  \\
% \midrule
% NILIF \citep{lei2025spike2former} & 80.98  & 66.3  \\
% NASN & 81.11  & 67.5  \\
% % \midrule
% % Ternary ASN & todo  & todo  \\
% \bottomrule
% \end{tabular}
% \end{small}
% \end{table}

\subsection{Ablation Study} 
In this part, we systematically compared the performance of the ILIF family (including ILIF and NILIF) and the ASN family (including ASN and NASN).

% \textbf{Ablation study on CIFAR and GLUE.} 
% In this part, we carry out the ablation study for the ILIF family (including ILIF and Normalized ILIF) and our ASN family (ASN and Normalized ASN) on CIFAR-100 and GLUE datasets. 
% %
% The results are presented in Table \ref{ablation_study}. ASN achieves superior performance over ILIF, while NASN outperforms NILIF. Consistent improvements are observed across both visual and language representation datasets, which further validates the effectiveness of our method.

\textbf{Ablation study on CIFAR dataset.} 
In this part, we carry out the ablation study for foundation spiking neurons on the CIFAR-100 dataset.
The results are presented in Table \ref{ablation_study}. Our NASN achieves the best performance. Compared to the original LIF-based model, ILIF, NILIF, ASN, and NASN can improve the performance. However, ASN achieves superior performance over ILIF, while NASN outperforms NILIF. Consistent improvements are observed, which further validate the effectiveness of our method.

\textbf{Neuron comparison on scaling.} 
In this part, we compare the performance differences between NILIF and our NASN during scaling. The experimental results show in Figure \ref{Neuron_scaling}. Notably, as scaling, the advantages of our method over NILIF become more pronounced, which further demonstrates that our method exhibits superior scaling properties compared to NILIF.

\section{Conclusion}
% In this work, we addressed the critical demand for high-performance and versatile spiking neurons in the era of large models. By establishing a novel functional perspective for spiking neuron design, we proposed the Adaptive Spiking Neuron (ASN) family, which features adaptive membrane dynamics and an integer-training and spike-driven inference paradigm. 
% %
% To accommodate diverse architectural needs, we further introduced Normalized variants to enhance training stability. Comprehensive evaluations across 19 datasets spanning vision and language modalities demonstrate the effectiveness and universality of our approach. We anticipate that the ASN family will serve as a foundational component for future large-scale spiking neural networks, bridging the gap between biological plausibility and modern AI performance to become the new generation for general-purpose spiking neurons.

In this work, we present a unified functional perspective for designing next-generation spiking neurons, emphasizing the joint importance of training efficiency, architectural compatibility, spike-driven inference, and adaptive firing dynamics. Motivated by the limitations of existing neuron models, we propose the Adaptive Spiking Neuron (ASN), which introduces learnable membrane potential dynamics to enable flexible and effective adaptive firing while maintaining computational efficiency through an integer training and spike inference paradigm. Furthermore, we develop the Normalized Adaptive Spiking Neuron (NASN) to enhance training stability via normalization.
Extensive experiments across 19 datasets spanning multiple vision and language tasks demonstrate the effectiveness and universality of our approach. Overall, this work provides both a practical neuron design and a general guideline for future research, and we believe the ASN family represents a promising step toward general-purpose spiking neurons for scalable and energy-efficient artificial intelligence.

\section*{Limitation}
Our energy consumption estimates are based on theoretical calculations and do not include measurements on real neuromorphic hardware. In addition, although we have demonstrated the effectiveness and generality of our method across 19 vision and language datasets with consistent results, further validation on neuromorphic tasks—such as DVS-based applications—remains an important direction for future work.

% In addition, regarding model scaling, our approach has been successfully extended to the 1B-parameter regime via direct training, where the advantages of our neuron become increasingly pronounced compared to the ILIF family. Nevertheless, a more systematic investigation of scaling behavior is required to assess whether emergent properties of spiking intelligence may arise at larger scales. These directions will leave for the next work in the future.

% \newpage

\bibliographystyle{ACM-Reference-Format}
\bibliography{sample-base}

\end{document}